\documentclass{article} 
\usepackage{iclr2025_conference}
\usepackage{times}
\usepackage{amsmath,amssymb,amsthm,graphicx,hyperref,xcolor,tikz,float,comment,booktabs,tcolorbox}
\usepackage{xcolor}
\definecolor{comp1}{HTML}{00B3B8}
\definecolor{comp2}{HTML}{ED135A}
\usepackage{amsmath}
\newenvironment{doublecase}
  {\left\lbrace\begin{aligned}}
  {\end{aligned}\right\rbrace}
\newenvironment{squaredcase}
  {\left[\begin{aligned}}
  {\end{aligned}\right]}
\usepackage{svg}

\usepackage{amsmath,amsfonts,bm}

\def\eqref#1{equation~\ref{#1}}

\def\1{\bm{1}}

\DeclareMathAlphabet{\mathsfit}{\encodingdefault}{\sfdefault}{m}{sl}
\SetMathAlphabet{\mathsfit}{bold}{\encodingdefault}{\sfdefault}{bx}{n}

\usepackage{url}

\usepackage[T1]{fontenc}    
\usepackage{url}            
\usepackage{booktabs}       
\usepackage{amsfonts}       
\usepackage{nicefrac}       
\usepackage{microtype} 
\newcommand{\methodnameNS}{Set-Based Prompting}
\usepackage{tikz}
\usetikzlibrary{positioning}
\usepackage{xcolor}
\usepackage{amsmath}
\usepackage{amssymb}
\usepackage{lmodern} 
\usepackage{caption}

\newcommand{\CheckedBox}{\text{\large\checkmark}}
\newcommand{\XBox}{\text{\large$\times$}}

\title{Order Independence With Finetuning}
\author{%
\begin{minipage}[t]{0.45\linewidth}
\raggedright
{\textbf{Katrina Brown}}\\[1ex]
{\normalfont Harvard University}\\[1ex]
{\normalfont \texttt{katrinabrown@college.harvard.edu}}
\end{minipage}
\hfill
\begin{minipage}[t]{0.45\linewidth}
\raggedright
{\textbf{Reid McIlroy}}\\[1ex]
{\normalfont Harvard University}\\[1ex]
{\normalfont \texttt{reidmcilroy@college.harvard.edu}}
\end{minipage}
}

\date{\today}
\begin{document}

\iclrfinalcopy

\maketitle

\begin{abstract}
Large language models (LLMs) demonstrate remarkable performance on many NLP tasks, yet often exhibit \emph{order dependence}: simply reordering semantically identical tokens (e.g., answer choices in multiple-choice questions) can lead to inconsistent predictions. Recent work proposes \emph{Set-Based Prompting} (SBP) as a way to remove order information from designated token subsets, thereby mitigating positional biases. However, applying SBP on base models induces an out-of-distribution input format, which can degrade in-distribution performance. We introduce a fine-tuning strategy that \emph{integrates} SBP into the training process, “pulling” these set-formatted prompts closer to the model’s training manifold. We show that SBP can be incorporated into a model via fine-tuning. Our experiments on in-distribution (MMLU) and out-of-distribution (CSQA, ARC Challenge) multiple-choice tasks show that SBP fine-tuning significantly improves accuracy and robustness to answer-order permutations, all while preserving broader language modeling capabilities. We discuss the broader implications of order-invariant modeling and outline future directions for building fairer, more consistent LLMs.
\end{abstract}

\section{Introduction}
\label{sec:intro}

Large language models (LLMs) based on Transformers \citep{vaswani2017attention,devlin2018bert} have achieved impressive zero-shot and few-shot performance across diverse NLP tasks \citep{brown2020language,radford2019language,touvron2023llama,touvron2023llama2}. Despite these advances, LLMs can be surprisingly sensitive to minor changes in input formatting. One notable instance of this \emph{order dependence} arises in multiple-choice question answering, where reordering semantically identical answer options can flip a model’s prediction \citep{talmor2019commonsenseqa,alzahrani_when_2024,zheng_large_2024}.

This vulnerability not only poses a practical challenge for building fair and reliable systems but also highlights lingering spurious correlations in LLMs’ learned representations. Figure~\ref{fig:contribution_redesign} illustrates a typical example with Llama-2, in which reversing the order of answer options changes the model’s response to a previously correct question.

One recent approach to mitigating order dependence is \emph{Set-Based Prompting (SBP)}, introduced by \citet{mcilroyyoung2024orderindependencefinetuning}. SBP reformats specified subsets of tokens (e.g., answer options) so that they receive no positional information, making the model’s output invariant to permutations of those subsets. However, applying SBP at inference time alone can degrade in-distribution performance. Because SBP prompts look unlike the sequences the model saw during pretraining, a distribution shift arises.

In this work, we propose to bridge this gap by fine-tuning LLMs on SBP-formatted data. Our core insight is that including SBP examples in the training regime “pulls” set-formatted prompts into the model’s learned manifold, reducing the mismatch that leads to performance drops. We adopt a margin-based contrastive loss that explicitly enforces separation between correct and incorrect answers. This choice addresses a key limitation of standard cross-entropy approaches, which maximize the probability of the correct option but do not strongly penalize near-ties with distractors.

\newpage
The key contributions of this work are as follows:
\begin{itemize}
    \item We demonstrate that fine-tuning with \methodnameNS{} formatted inputs significantly improves the order-independent \methodnameNS{} question-answering accuracy, addressing the issue of performance degradation observed in \cite{mcilroyyoung2024orderindependencefinetuning}, and that these benefits generalize well to novel inputs.
    \item We analyze the practical best methods in finetuning to elicit these performance gains. In particular, we show that margin-based contrastive training significantly outperforms standard cross-entropy in aligning \methodnameNS{} prompts with the model’s decision boundary.
\end{itemize}

We close with a discussion of potential applications for Set-Based Prompting (SBP) based approaches in other tasks (e.g., pairwise ranking, summarization) and highlight ongoing challenges in building fully order-invariant NLP systems.

\section{Related Works}
\label{sec:related}
The Transformer architecture~\citep{vaswani2017attention} underpins a range of LLMs~\citep{devlin2018bert,brown2020language,touvron2023llama,touvron2023llama2} that excel in summarization, question answering, and more. Yet recent studies note that even large models can falter with long or perturbed inputs \citep{liu2024lost}.

\subsection{Order Dependence and Prompt Sensitivity.}
Multiple-choice QA is particularly prone to positional biases: reversing or permuting the answer candidates can yield divergent results \citep{talmor2019commonsenseqa,alzahrani_when_2024,zheng_large_2024}. Researchers have also found similar vulnerabilities in pairwise comparison tasks \citep{liusie2024comparative} and used positional “tells” to detect training-data contamination \citep{oren2023proving}. Such observations highlight the need for strategies that make models \emph{invariant} to superficial reordering.

\subsection{Set-Based Prompting (SBP).}
\begin{figure}
\begin{minipage}[t]{0.33\textwidth}
  \begin{minipage}{0.9\textwidth}
    \begin{tcolorbox}[
      title={a) Llama 7B - Base}\\{\small\textit{Default Ordering}},
      colbacktitle=comp1, colback=comp1!8!white, 
      height=5.5cm, valign=center
    ]
      Answer the following question:\:\includegraphics[height=2ex]{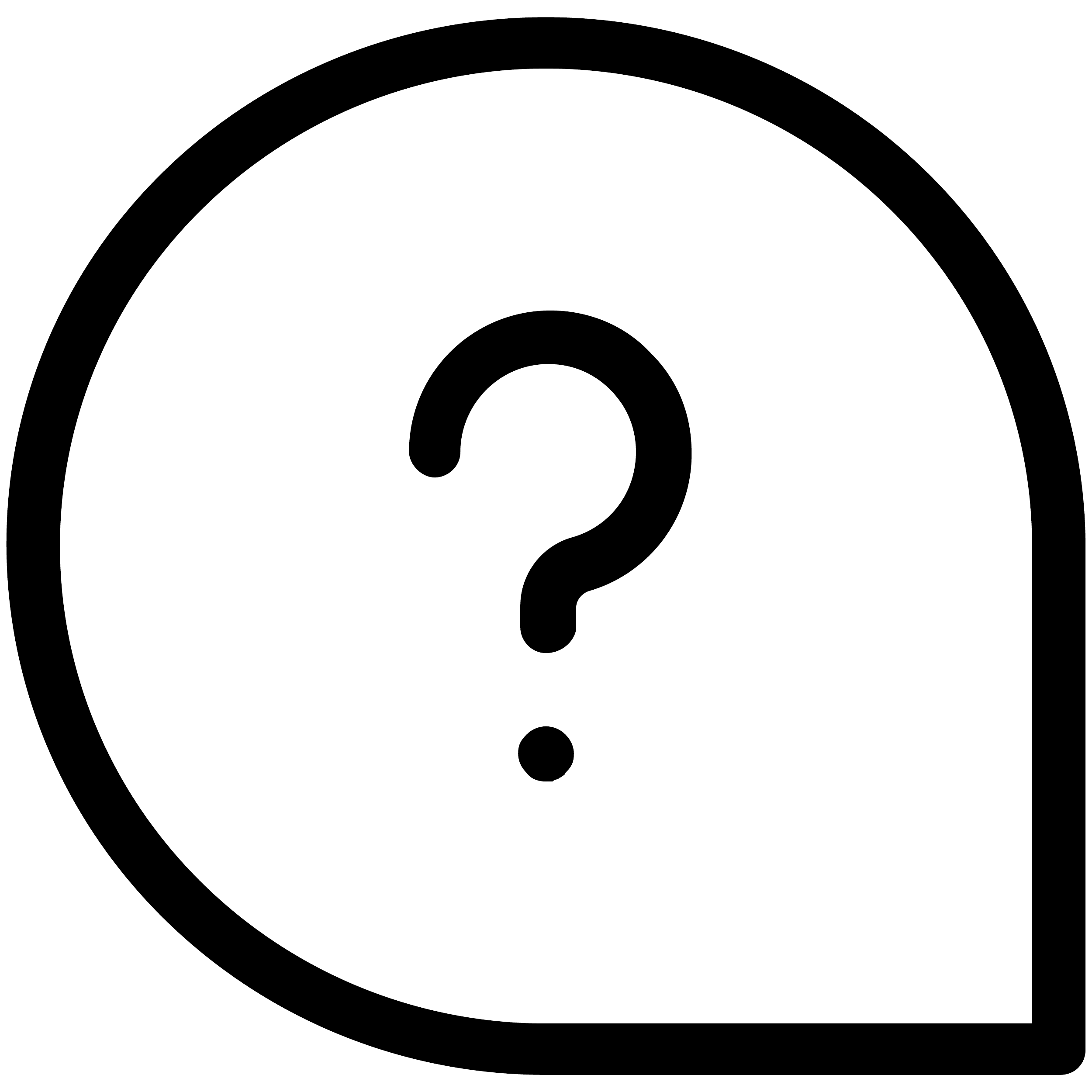}
      \begin{align*}
        \begin{squaredcase}
          \quad \includegraphics[height=2ex]{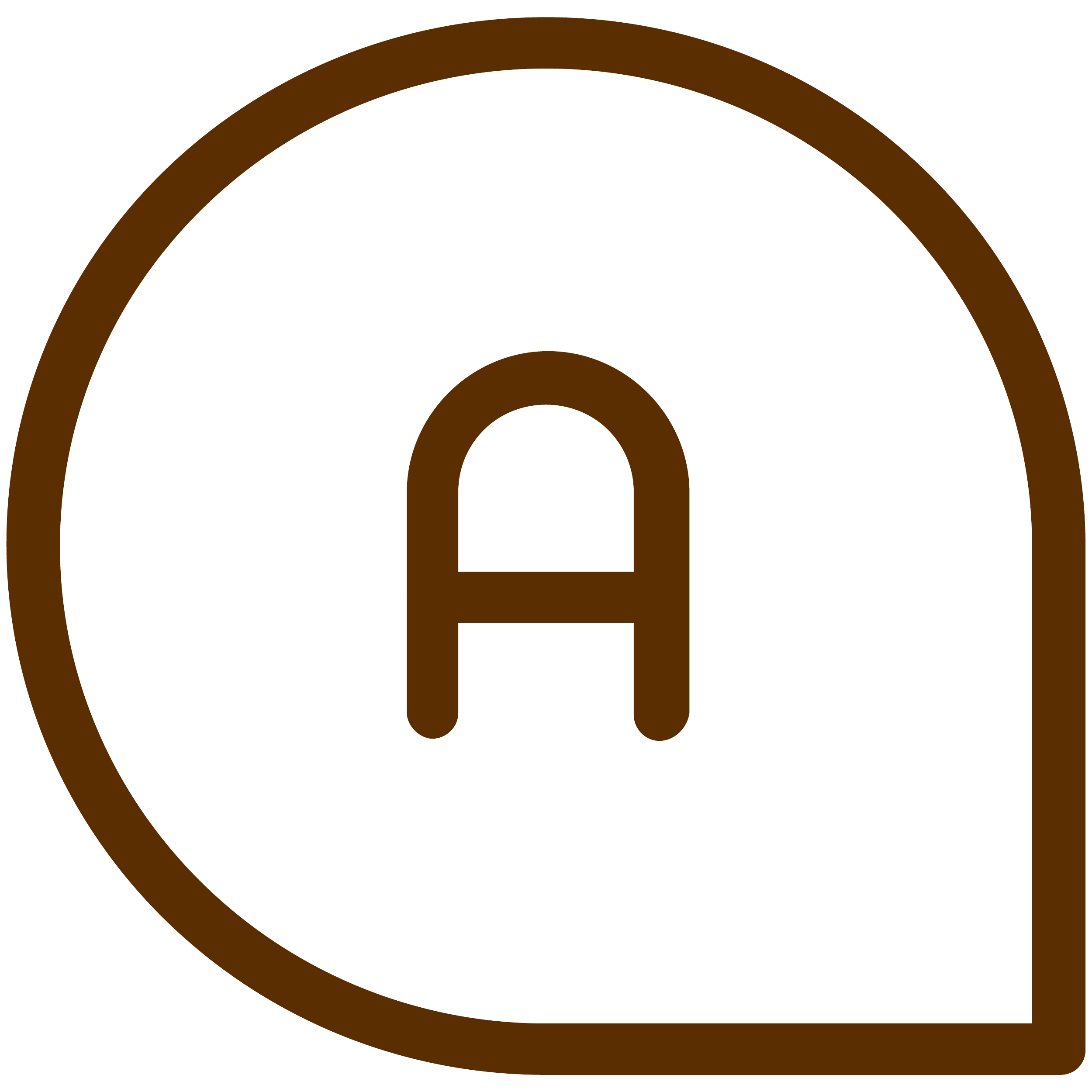},\quad \\
          \quad \includegraphics[height=2ex]{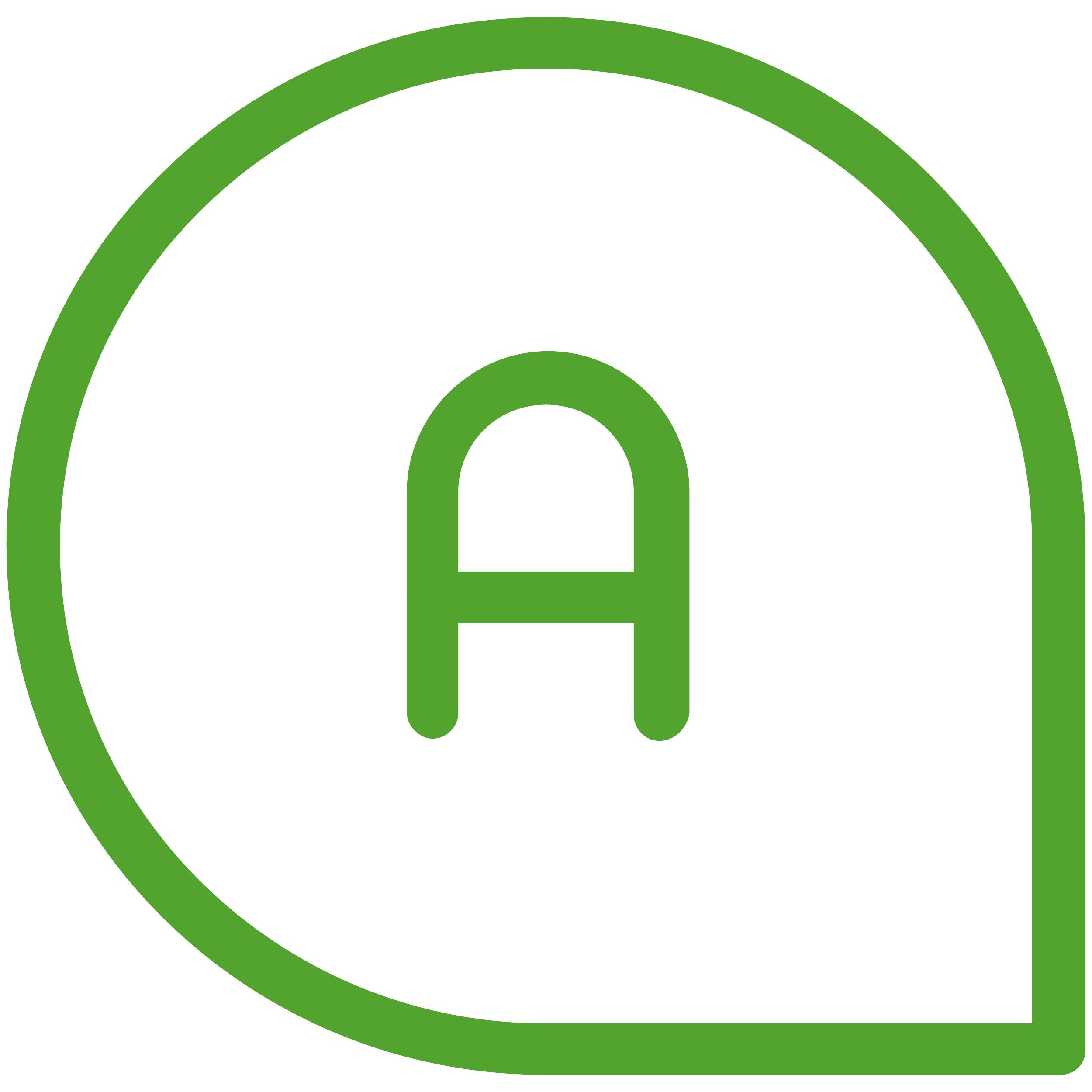},\quad \\
          \quad \includegraphics[height=2ex]{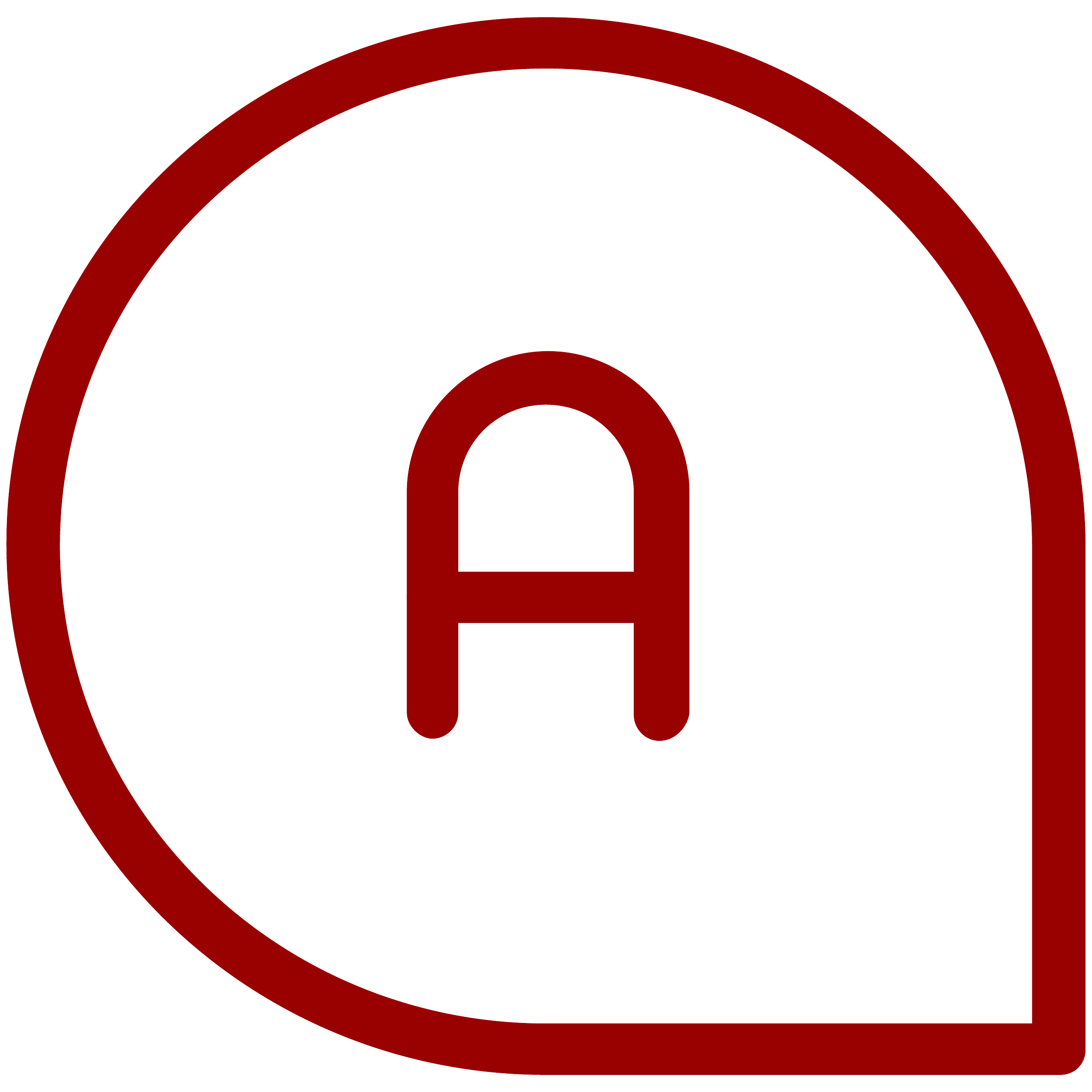},\quad \\
          \quad \includegraphics[height=2ex]{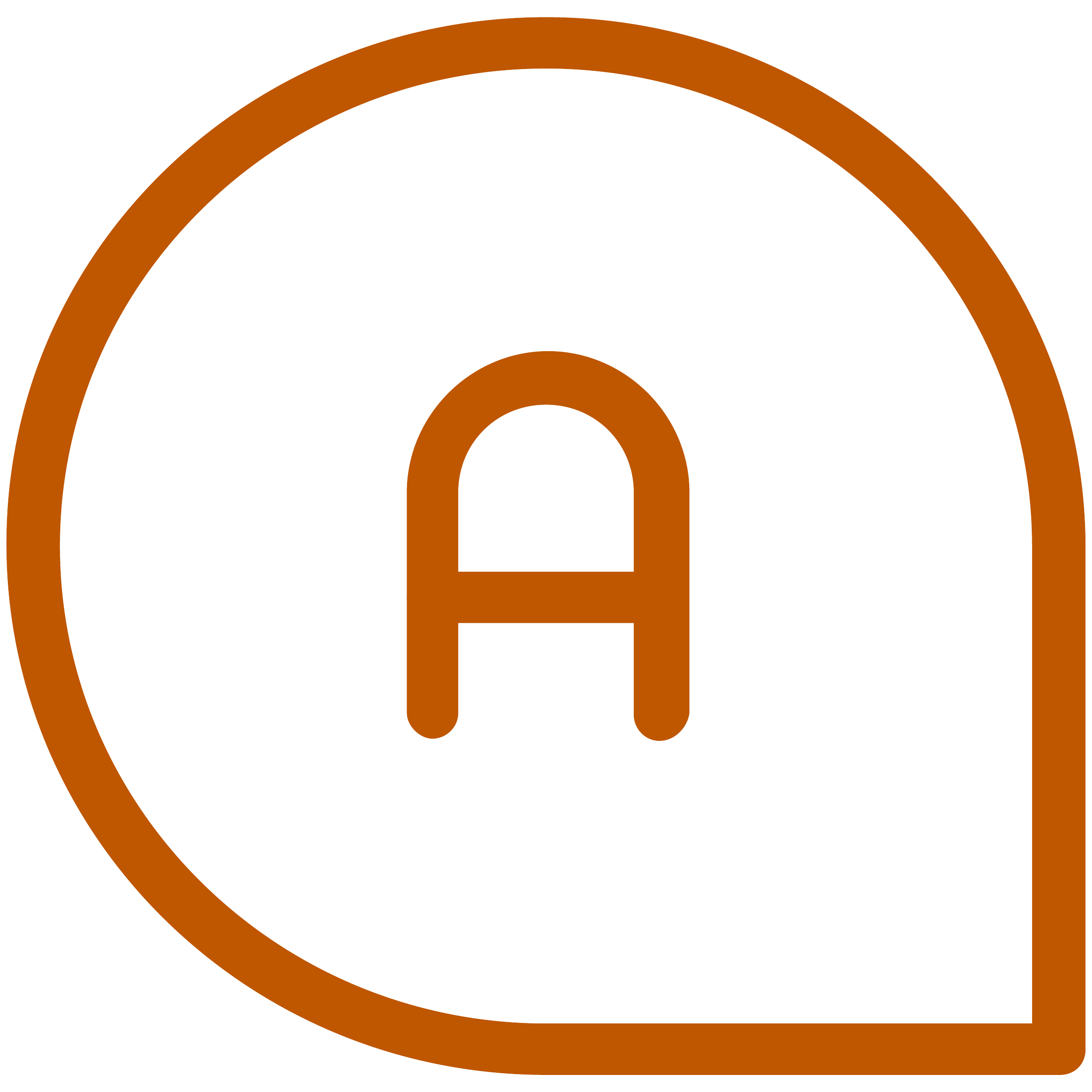},\quad \\
        \end{squaredcase}
      \end{align*}
      \tcblower
      Answer: \quad \includegraphics[height=2ex]{images/noun-answer-7151735-green1.pdf} \quad \CheckedBox
    \end{tcolorbox}
  \end{minipage}
\end{minipage}%
\hfill
\begin{minipage}[t]{0.33\textwidth}
  \begin{minipage}{0.9\textwidth}
    \begin{tcolorbox}[
      title={b) Llama 7B - Base}\\{\small\textit{Reversed Order}},
      colbacktitle=comp2, colback=comp2!8!white, 
      height=5.5cm, valign=center
    ]
      Answer the following question:\:\includegraphics[height=2ex]{images/noun-question-7151715.pdf}
      \begin{align*}
        \begin{squaredcase}
          \quad \includegraphics[height=2ex]{images/noun-answer-7151735-red3.pdf},\quad \\
          \quad \includegraphics[height=2ex]{images/noun-answer-7151735-red2.pdf},\quad \\
          \quad \includegraphics[height=2ex]{images/noun-answer-7151735-green1.pdf},\quad \\
          \quad \includegraphics[height=2ex]{images/noun-answer-7151735-red1.pdf},\quad \\
        \end{squaredcase}
      \end{align*}
      \tcblower
      Answer:\: \includegraphics[height=2ex]{images/noun-answer-7151735-red3.pdf} \quad \XBox
    \end{tcolorbox}
  \end{minipage}
\end{minipage}%
\hfill
\begin{minipage}[t]{0.33\textwidth}
  \begin{minipage}{0.9\textwidth}
    \begin{tcolorbox}[
      title={c) Llama 7B - Base}\\{\small\textit{Set-Based Prompting}},
      colbacktitle=comp1, colback=comp1!8!white, 
      height=5.5cm, valign=center
    ]
      Answer the following question:\:\includegraphics[height=2ex]{images/noun-question-7151715.pdf}
      \begin{align*}
        \begin{doublecase}
          \quad \includegraphics[height=2ex]{images/noun-answer-7151735-red3.pdf},\quad \\
          \quad \includegraphics[height=2ex]{images/noun-answer-7151735-red2.pdf},\quad \\
          \quad \includegraphics[height=2ex]{images/noun-answer-7151735-green1.pdf},\quad \\
          \quad \includegraphics[height=2ex]{images/noun-answer-7151735-red1.pdf},\quad \\
        \end{doublecase}
      \end{align*}
      \tcblower
      Answer: \: \includegraphics[height=2ex]{images/noun-answer-7151735-green1.pdf} \quad \CheckedBox
    \end{tcolorbox}
  \end{minipage}
\end{minipage}
\caption{Visualization of order dependency in Llama 2, 7B, when asked to choose the best among three resumes. In variant (a) the default ordering leads to a correct answer. Variant (b) reverses the answer choices and results in an incorrect response, while variant (c) applies \methodnameNS{} to neutralize ordering effects, restoring the correct answer.}
\label{fig:contribution_redesign}
\end{figure}

\citet{mcilroyyoung2024orderindependencefinetuning} propose a method to \emph{remove} positional signals for subsets of tokens. Specifically, as visualized in Figure~\ref{fig:contribution_redesign}, SBP applies (1) modified attention masks that do not enforce strict left-to-right order within certain sub-sequences, and (2) identical or parallel positional embeddings for tokens in that sub-sequence. As discussed above, SBP can yield order-invariant predictions for multiple-choice tasks. Nonetheless, applying SBP to a model that has never seen such prompts (during training) induces an out-of-distribution mismatch, potentially causing performance dips on standard queries. Our work addresses this limitation by explicitly fine-tuning on SBP data.

\subsection{Instruction Tuning and Fine-Tuning Strategies.}
Instruction tuning \citep{ouyang2022training,wang2022self} guides a model to follow user intents more closely, while parameter-efficient methods like LoRA~\citep{peft} allow specialized fine-tuning of large models. We adopt such techniques for SBP integration, using a margin-based contrastive objective \citep{gunel2021supervisedcontrastivelearningpretrained} that better separates correct from incorrect answers.

\subsection{Contrastive Objectives in Multiple-Choice QA and Beyond.} Contrastive learning has emerged as a powerful framework for both supervised and self-supervised tasks \citep{chen2020simpleframeworkcontrastivelearning,oord2019representationlearningcontrastivepredictive,chuang2020debiasedcontrastivelearning}. In broad terms, these methods aim to pull semantically similar embeddings closer while pushing dissimilar ones apart. Within multiple-choice QA, researchers have explored various contrastive strategies to emphasize the gap between correct and incorrect choices. For instance, \citet{yao2021contextguidedtriplematchingmultiple} introduce a context-guided triple matching method that applies contrastive regularization to distinguish the correct answer from distractors. Although these approaches often embed additional context or perform complex matching across passage, question, and answer, they align with our motivation to enforce clearer separation of logits or embeddings for correct vs.\ incorrect candidates.

Compared to these prior works, our margin-based loss similarly promotes a separation between ground-truth and distractor answers but is integrated into Set-Based Prompting and fine-tuning on LLMs. In particular, we apply contrastive signals specifically to realign the model when an SBP format removes the usual positional cues. By adopting methods inspired by self-supervised contrastive research \citep{chen2020simpleframeworkcontrastivelearning,chuang2020debiasedcontrastivelearning}, we ensure that SBP does not degrade performance on in-distribution tasks and remains robust to reordering. Notably, whereas prior contrastive QA methods \citep{yao2021contextguidedtriplematchingmultiple} typically focus on triple matching or more intricate alignment, our approach simplifies the problem by structurally removing order information, thus reducing the likelihood of position-based biases and reinforcing the contrastive boundary through margin-based separation.

\subsection{Evaluation Benchmarks and Robustness.}
Datasets like MMLU \citep{hendrycks2021measuringmassivemultitasklanguage}, CommonsenseQA \citep{talmor2019commonsenseqa}, and ARC \citep{clark2018thinksolvedquestionanswering} stress the reasoning abilities of LLMs under standard multiple-choice formats. Recent work also explores how subtle prompt edits can cause large swings in performance \citep{alzahrani_when_2024,zheng_large_2024}. Order dependence has also been observed on information retrieval tasks, via the `lost-in-the-middle' effect~\citep{liu2024lost}. Our method systematically addresses such vulnerabilities by exposing the model to SBP-style prompts during training, thereby producing consistency across permutations.

\section{Experimental Procedure}
\label{sec:experiments}

In this section, we detail our experimental setup designed to evaluate the efficacy of fine-tuning large language models (LLMs) on Set-Based Prompting (SBP) data. We assess whether SBP fine-tuning can effectively bring SBP-formatted inputs closer to the model’s training manifold, providing robustness to input order permutations without compromising performance. We conduct experiments on the MMLU dataset for finetuning, and evaluate generalization using CSQA and ARC Challenge. In addition, we monitor WikiText-103 perplexity \citep{merity2016pointer} to ensure that our approach does not degrade the model’s broader language modeling capabilities.

\subsection{Datasets}
We employ three distinct multiple-choice question (MCQ) benchmarks: the MMLU benchmark \citep{hendrycks2021measuring} (4 questions), CommonsenseQA (CSQA) \citep{talmor2019commonsenseqa} (5 questions), and ARC Challenge \citep{clark2018thinksolvedquestionanswering} (4 questions\footnote{Questions with under 4 answers were removed}). We preprocess the data by filtering questions so that the tokenized question–answer pairs do not exceed 256 tokens and contain at least three incorrect answers. This yields 12,147 MMLU questions, 9,741 CSQA questions, and 2,582 ARC Challenge questions. Following the original \methodnameNS{} approach \citep{mcilroyyoung2024orderindependencefinetuning}, we convert numeric or alphabetic labels into quoted text snippets (e.g., \texttt{"optionA"}, \texttt{"optionB"}) to ensure consistency when transforming answer options into parallel sub-sequences.\\
We finetune the model only on data from MMLU, while we evaluate question answer accuracy on all three MCQ benchmarks. In practice, this means that the accuracy as reported on MMLU is "in-distribution" train accuracy, since the model was finetuned on this data, while the accuracy as reported on the other two benchmarks is "out-of-distribution" test accuracy, since this data is unseen by the model during the fine-tuning stage. We measure both question answering accuracy under \methodnameNS{} as well as question answer accuracy under standard order dependent prompting. For the latter, we measure the accuracy under order dependent prompting for all permutations of the answer options (e.g. QA accuracy when answer options in the question statement are reversed), yielding a measure of the model's order sensitivity under permutation. To compute which MCQ option is selected as `correct' by the model, for each candidate option, we compute the average log-probability of its tokens (conditioned on the question) and select the option with the highest score.

\subsubsection{WikiText-103 (Monitoring Language Modeling Capabilities)}
In addition to MCQ performance, we track perplexity on WikiText-103 \citep{merity2016pointer} to verify that SBP fine-tuning does not significantly impair the model’s general language modeling ability. A marked increase in perplexity would indicate that the model’s core generative aptitude has been compromised by SBP finetuning.

\subsection{MCQ Interventions and Baselines}
\methodnameNS{} Fine-Tuning (Treatment): 
Our primary intervention involves fine-tuning each LLM on the MMLU dataset with answer options reformatted \methodnameNS{} parallel sub-sequence structure. The objective is to bring these SBP inputs closer to the model’s training manifold, thereby improving robustness to permutations and enhancing order invariance.

Standard Fine-Tuning (Control): 
For comparison, we fine-tune each model on MMLU data using the standard, order-dependent format (i.e., without \methodnameNS{} formatting). This baseline allows us to isolate the accuracy gains attributable to finetuning on MCQ questions in general from the accuracy gains attributable specifically to finetuning on \methodnameNS{} data.

No Fine-Tuning Baseline (Base): 
We also evaluate the base models (without additional fine-tuning) as a zero-shot baseline, which enables us to gauge the performance shift resulting from both SBP and standard fine-tuning.

\subsubsection{Base Models}
We experiment with two variants of LLaMA-2~\citep{touvron2023llama2}: a base model (\texttt{Llama-2-7b}) and an instruction-tuned model (\texttt{Llama-2-7b-chat}). Table~\ref{tab:llmstable} lists the model details. 

\subsection{Choice of Loss Function}

We examine the impact of two loss functions during fine-tuning:\\\\
Standard Cross-Entropy Loss: Compute the negative average log probability of tokens in the answer sequence conditional on the question tokens. The standard cross-entropy loss for a token sequence $\mathbf{x} = (x_1, x_2, \dots, x_T)$ is given by $L = -\frac{1}{T} \sum_{t=1}^{T} \log p_\theta(x_t \mid x_{<t})$, 
where $x_{<t}$ represents the preceding tokens and $\theta$ denotes the model parameters.\\\\
Margin Based Contrastive Loss: Compute the average per-token log-probability $p$ of the correct answer sequence (conditioned on the question) and likewise $\{n_1, n_2, \dots, n_k\}$ for the $k$ incorrect answer sequences (conditioned on the question), with the loss defined as:
    \[
    L = \max(0, m - (p - \max(n_1, n_2, \dots, n_k))).
    \]
    This yields a differentiable objective that refines the model’s decision boundary by increasing the probability of generating the correct answer tokens while decreasing the probability of generating the incorrect answer tokens.

\subsubsection{Finetuning Methodology}
For optimization, we use the AdamW optimizer~\citep{loshchilov2017decoupled}. To reduce the computational burden of fine-tuning, we adopt the LoRA approach for parameter-efficient tuning using the PEFT framework \citep{peft}. Figure~\ref{fig:loss} shows the training and validation loss curves, which demonstrate stable convergence without signs of overfitting.

\section{Results}
\label{sec:results}

Below we present our experimental results to evaluate the impact of SBP fine-tuning on both in-distribution and out-of-distribution multiple-choice question answering (MCQ) tasks, as well as on general language modeling via WikiText-103 perplexity. Our experiments compare models fine-tuned with \methodnameNS{}-formatted data (treatment) against those fine-tuned using standard, order-dependent prompts (control) and against the original base models. We compare results from finetuning under either the contrastive loss function or the standard cross-entropy loss function.

\begin{figure}[t]
\centering
\begin{minipage}{\linewidth}
    \centering
    \includegraphics[width=\textwidth]{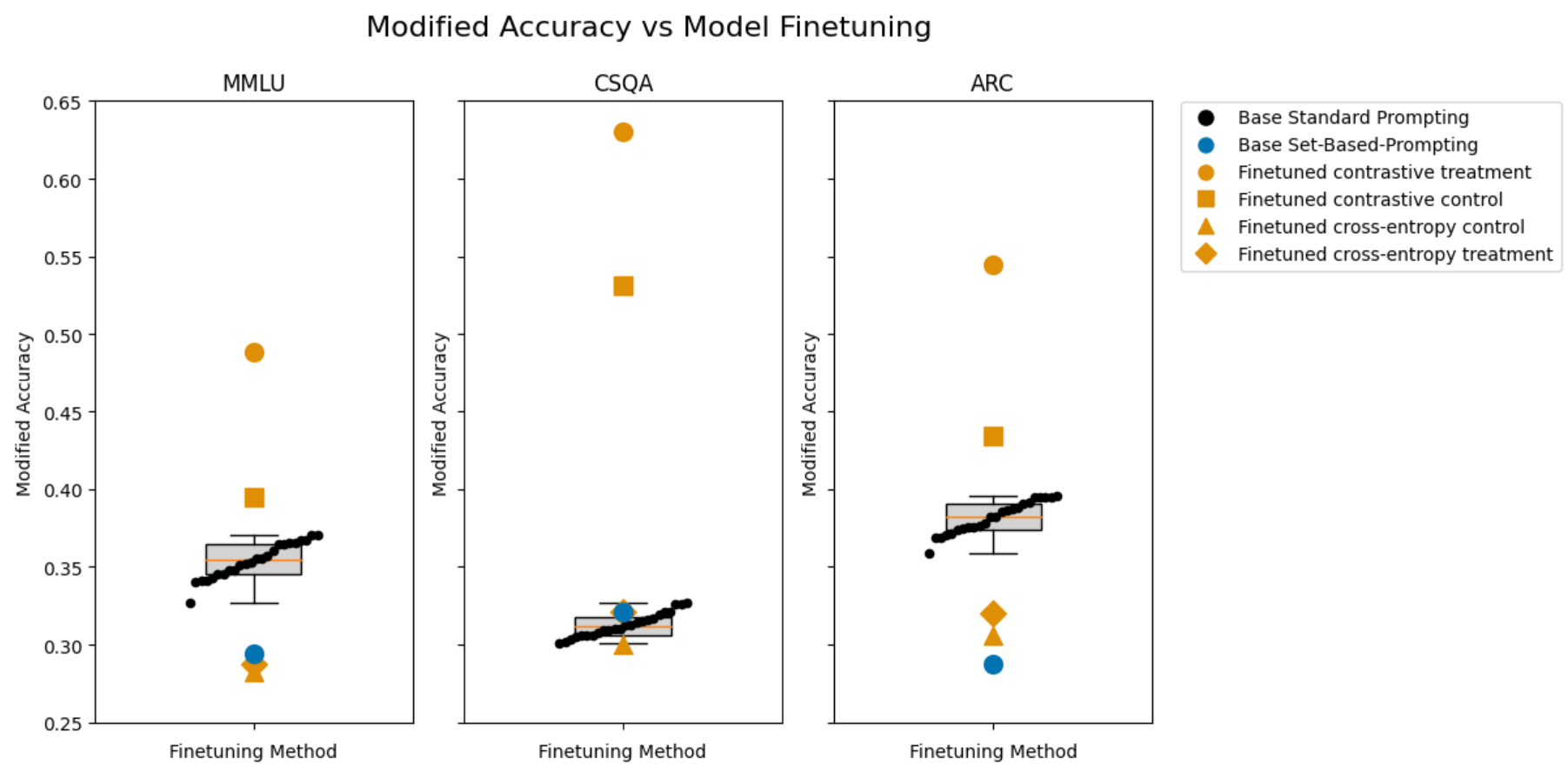}
    \caption{Question answering accuracy for \texttt{Llama-2-7b-hf} under 4! reorderings for Standard Prompting on the base model, and for \methodnameNS{} on the base and finetuned models. Note on the legend that contrastive vs cross-entropy indicates the loss function used during finetuning to obtain the finetuned model, while treatment vs control indicates whether the model was finetuned on \methodnameNS{} (treatment) vs standard order dependent (control) formatted data.}
    \label{fig:boxplot_7b}
\end{minipage}
\end{figure}

\subsection{Robustness to Input Permutations}

Figure~\ref{fig:boxplot_7b} illustrates the performance of the \texttt{Llama-2-7b} model under 24 ($4!$) different reorderings of answer options. In the base models, accuracy varies considerably under reordering of the answer options, underscoring a strong order dependency. Across all datasets, Finetuned \methodnameNS{} QA accuracy is signicantly higher when the model was finetuned with \methodnameNS{} (treatment) data than with standard order dependent (control) data. Likewise across all datasets, the Finetuned \methodnameNS{} QA accuracy is significantly higher when finetuning under a contrastive loss function than under the standard cross-entropy loss function (note that QA accuracy actually decreases under the standard cross-entropy loss function, signifying misalignment between the loss function and the QA objective). Figure ~\ref{fig:boxplot_7b_chat} illustrates that similar effects hold for \texttt{Llama-2-7b-chat}.

\subsection{Impact of Loss Functions}
\label{sec:loss-comparison}
One key finding from our experiments is that the choice of loss function significantly influences how well the model adapts to \methodnameNS{}-formatted inputs. We compare two approaches: (1) a standard cross-entropy loss applied only to the answer tokens, and (2) a margin-based contrastive loss that enforces a separation between correct and incorrect answers. 

In principle, standard cross-entropy encourages high probability for the correct answer. However, it does not explicitly penalize the model if a distractor (incorrect) option is scored nearly as high. Consequently, the model may focus too narrowly on maximizing the probability of the correct sequence without robustly separating it from the incorrect sequences in logit space. In our experiments, this lack of explicit separation manifests as deteriorating performance across all datasets. 

In contrast, the margin-based contrastive  loss aims to explicitly push the model to not only boost the probability of the correct answer but also demote the probabilities of all incorrect answers by a certain margin $m > 0$. 

We observe that adopting margin based contrastive loss consistently yields higher accuracy under SBP prompts while also improving robustness to answer reordering. The margin-based loss produces a tighter alignment between the model’s confidence (log-probabilities) and correctness, ultimately leading to stronger calibration.

The results indicate that the improvements from SBP fine-tuning are not solely attributable to exposure to an augmented prompt format. Rather, they stem from effective calibration of the model’s logits, enforced by the contrastive margin. In other words, when the model is trained to maintain a non-trivial gap between correct and incorrect answers, it learns a more robust internal representation of the answer space. 

By comparing the two loss functions, we conclude that margin-based contrastive training is key to achieving high performance under SBP prompts. 

\subsection{Best-of/Worst-of Performance}
\label{sec:bestof_worstof}

We measure the model's sensitivity to small changes in the presentation of answer options using three metrics across two permutations of each multiple-choice question, namely a normal ordering versus a reversed ordering. The first metric, Best-of-2 Accuracy, is the fraction of questions for which the model produces a correct answer under at least one ordering. That is,
\[
\text{Best-of-2} = \frac{\bigl|\bigl\{\,q : \text{Correct}(\text{normal}, q)\,\lor\,\text{Correct}(\text{reversed}, q)\bigr\}\bigr|}{\text{Total Number of Questions}},
\]
where Correct(\(\cdot\), $q$) indicates that the model selected the correct option for question $q$ under the specified ordering. The second metric, Best-of-1 Accuracy, is the fraction of questions for which the model is correct only under the normal ordering; this serves as a baseline measure of single-prompt performance. The final metric, Worst-of-1 Accuracy, is the fraction of questions for which the model is incorrect under at least one ordering, indicating how prone the model is to making mistakes whenever the ordering deviates from what it expects. A large gap between Best-of-2 and Worst-of-1 implies high sensitivity to prompt format, whereas a smaller gap suggests greater robustness.

\begin{figure}[t]
\begin{minipage}{1.0\linewidth}
    \centering
    \includegraphics[width=\textwidth]{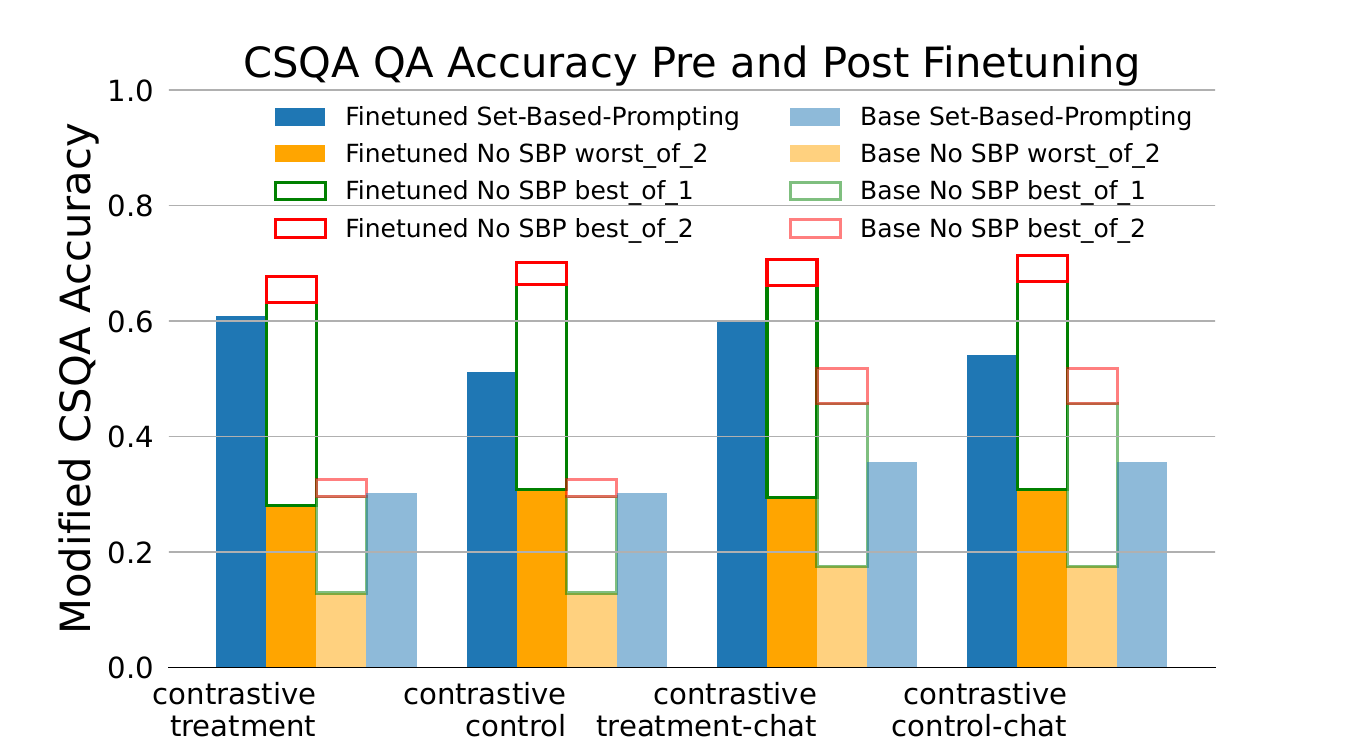}
    \caption{On CSQA questions (which were unseen in the data used for finetuning), \methodnameNS{} accuracy post-fine-tuning significantly exceeds pre-fine-tuning best-of-2 accuracy. Note on the x-axis that the \texttt{chat} suffix indicates testing on \texttt{Llama-2-7b-chat} while the absence of this suffix indicates testing on \texttt{Llama-2-7b}.}
    \label{fig:barplot_datasets}
\end{minipage}
\end{figure}

In the base model, the order independent accuracy in the instruct-tuned model is significantly below that of the Best-of-1 normal accuracy, the concern raised in \citet{mcilroyyoung2024orderindependencefinetuning} that \methodnameNS{} degrades task performance. Figure~\ref{fig:barplot_datasets} demonstrates that after fine-tuning with \methodnameNS{} formatted data under a contrastive loss function (contrastive treatment), both the Llama-2-7b and Llama-2-7b-chat models under order independent \methodnameNS{} surpass their own Best-of-2 accuracy levels from before finetuning, which demonstrates that \methodnameNS{} training significantly improves overall output quality.  Figure~\ref{fig:barplot_mmlu_arc} shows that similar improvements generalize to the MMLU and ARC benchmarks, suggesting that explicitly neutralizing positional cues provides a reliable path toward more robust multiple-choice question answering.

\subsection{Non-Question Answering Performance}

To ensure that \methodnameNS{} fine-tuning does not compromise the model's general language modeling capabilities, we monitor perplexity on WikiText-103 \citep{merity2016pointer}. Table~\ref{tab:perplexity} shows the initial and final perplexity for both \methodnameNS{} and standard fine-tuning across the two LLaMA-2 variants, when finetuning on either \methodnameNS{} formatted data (treatment) or standard formatted data (control). For \texttt{Llama-2-7b}, the perplexity increases marginally from 12.66 to 12.76 (treatment) and to 12.81 (control). In contrast, for the instruction-tuned \texttt{Llama-2-7b-chat}, perplexity decreases from approximately 17.04 to 15.36 (treatment) and to 15.85 (control). These results indicate that SBP fine-tuning does not adversely affect the model's underlying language modeling performance, although it may "undo" some of the instruct tuning on the instruct model variant.

\begin{table}[t]
    \centering
    \caption{Perplexity Comparison Across Models and Fine-Tuning Formats}
    \label{tab:perplexity}
    \begin{tabular}{llcc}
        \toprule
        \textbf{Model} & \textbf{Data Type} & \textbf{Base Perplexity} & \textbf{Finetuned Perplexity} \\
        \midrule
        7b      & treatment & 12.66 & 12.76 \\
        7b      & control & 12.66 & 12.81 \\
        7b-chat & treatment & 17.04 & 15.36 \\
        7b-chat & control & 17.03 & 15.85 \\
        \bottomrule
    \end{tabular}
\end{table}

\subsection{Discussion and Limitations}
\label{sec:discussion-limitations}

Our experiments consistently show that Set-Based Prompting (SBP) fine-tuning provides a way to eliminate ordering bias while maintaining and improving performance in multiple-choice question answering. In both \texttt{Llama-2-7b} and \texttt{Llama-2-7b-chat}, SBP yields higher accuracy across in-distribution (MMLU) and out-of-distribution (CSQA, ARC Challenge) datasets compared to the base models or models finetuned on standard formatted data. In particular, finetuning eliminates and reverses the degradation in question-answering accuracy observed in \citep{mcilroyyoung2024orderindependencefinetuning} under Set-Based Prompting, highlighting the effectiveness of aligning SBP inputs with the model’s training manifold. The positive results on CSQA and ARC Challenge suggest that treating answer choices as sets helps the model avoid spurious positional cues, ultimately improving out-of-distribution performance. 

A comparison of loss functions indicates that a margin-based contrastive objective aids in maximizing these gains. By enforcing a margin between the correct answer’s log-probability and that of distractors, this objective prevents near-ties and encourages the model to rely less on superficial ordering. Despite these promising outcomes, several caveats remain. One pertains to applicability beyond multiple-choice question answering. Although SBP is conceptually extendable to other tasks such as ranking or summarization, this work focuses primarily on multiple-choice QA, and further experimentation is needed to confirm broader utility. Another consideration involves mixing SBP prompts with instruction-tuned data, which can slightly alter perplexity and potentially overwrite certain instruction-following behaviors, as suggested by the decrease in perplexity on \texttt{Llama-2-7b-chat}. Future research could explore mixing SBP formatted data into the instruct finetuning process to preserve desired conversational traits. A further limitation is that the margin-based loss is fixed at 1.0, and different tasks, model sizes, or data regimes may benefit from alternative margins or more nuanced loss formulations. Finally, this work fine-tunes on MMLU, which may not reflect the full diversity of question-answer distributions found in other domains; training on larger or more varied corpora could improve robustness.

\subsection{Summarization Task}
We evaluated whether a model finetuned under contrastive loss on \methodnameNS{} formatted inputs would have improved performance on additional tasks, such as summarization. We extracted excerpts of 20 sentences each from a dataset of reports on SEC filings \citep{financial-reports-sec}, split them into four groups of five sentences each, and prompted both the base and finetuned \texttt{Llama-2-7b-chat} models to summarize the excerpts under either standard order dependent prompting or \methodnameNS{}. Qualitatively, under both base and finetuned models, the quality of the summary produced under \methodnameNS{} was significantly worse (see Appendix section~\ref{sec:summary_outputs}) than the quality of the summary produced under standard order dependent prompting for the base model. Finetuning did not mitigate this degradation in summary quality, suggesting that the finetuning objective is misaligned with the objective of increasing \methodnameNS{} summary quality. However, the success of finetuning on MCQ accuracy under a task specific aligned objective with increasing QA accuracy under \methodnameNS{} motivates future study of finetuning under \methodnameNS{} formatted summarization inputs with a summarization task-aligned objective, towards order independent summaries that do not suffer from the `lost in the middle` effect observed in \citet{liu2024lost}.

\section{Conclusion}
\label{sec:conclusion}
Set-Based Prompting (SBP) fine-tuning offers a compelling framework for mitigating the well-documented sensitivity of large language models to token order in multiple-choice questions. By training directly on SBP-formatted examples with a margin-based contrastive objective, our approach guarantees that the correct option is consistently assigned a higher probability than distractors, effectively eliminating error tied to superficial variations in answer ordering. Our experiments suggest that these gains generalize well to unseen data, providing robustness to input permutations without sacrificing performance.

Moreover, the SBP pipeline is straightforward to incorporate with standard parameter-efficient finetuning techniques and can be adapted to a variety of tasks that rely on comparing or ranking text segments. We envision immediate applications in fairer assessment tools, where the order of presented answers should not affect outcomes. Looking ahead, extending SBP to more complex structured inputs could uncover additional benefits in domains such as recommender systems and structured summarization.

By demonstrating how contrastive training can fuse set-based invariance into large-scale language modeling, we provide both practical tools and conceptual insights for building more consistent and equitable NLP systems. These findings motivate further exploration of order invariance as a means of exposing—and ultimately alleviating—longstanding biases in base models.


\bibliography{iclr2025_conference}

\appendix
\section{Model Details}
\begin{table}[H]
    \centering
    \caption{Models used in this analysis}
    \label{tab:llmstable}
    \begin{tabular}{llccl}
    \toprule
    \textbf{Organization} & \textbf{Model Name} & \textbf{Parameters (B)} & \textbf{Instruction-Tuned?} & \textbf{Links}\\
    \midrule
    Meta & \texttt{Llama-2-7b} & 7 & No & \href{https://huggingface.co/meta-llama/Llama-2-7b-hf}{(HuggingFace)}\\
    Meta & \texttt{Llama-2-7b-chat-hf} & 7 & Yes & \href{https://huggingface.co/meta-llama/Llama-2-7b-chat-hf}{(HuggingFace)}\\
    \bottomrule
    \end{tabular}
\end{table}

\section{Finetuning Details}
All fine-tuning experiments were conducted on 4~$\times$~H100 GPUs. We use a batch size of 4, and employ a linear learning rate schedule with an initial learning rate of $2 \times 10^{-5}$. A warmup phase covering the first 10\% of training steps is applied, after which the learning rate decays linearly to zero. Formally, the learning rate at step $t$ is defined as:
\begin{equation}
    \text{lr}(t) = 
    \begin{cases}
        \alpha \cdot \frac{t}{w_{\text{steps}}}, & \text{if } t \leq w_{\text{steps}},\\[3pt]
        \alpha \cdot \frac{T - t}{T - w_{\text{steps}}}, & \text{if } t > w_{\text{steps}},
    \end{cases}
\end{equation}
where $\alpha = 2 \times 10^{-5}$, $T$ is the total number of training steps, and $w_{\text{steps}} = 0.1T$. We use the default AdamW hyperparameters: $\beta_1 = 0.9$, $\beta_2 = 0.999$, and $\epsilon = 10^{-8}$. All models are fine-tuned for exactly 3 epochs without exhaustive hyperparameter tuning or early stopping.

For parameter-efficient tuning, we adopt the LoRA framework \citep{peft} with rank 8, scaling factor ($\alpha$) 16, applied to the q, k, v, and o projections, LoRA dropout 5\%, no additional bias parameters, for causal language modeling. These hyperparameters were chosen as defaults and were not tuned.

\section{Additional Dataset Results}
Trends are similar across \texttt{Llama-2-7b-hf} and \texttt{Llama-2-7b-chat-hf}. 
\begin{figure}[h]
\begin{minipage}{\linewidth}
    \centering
    \includegraphics[width=\textwidth]{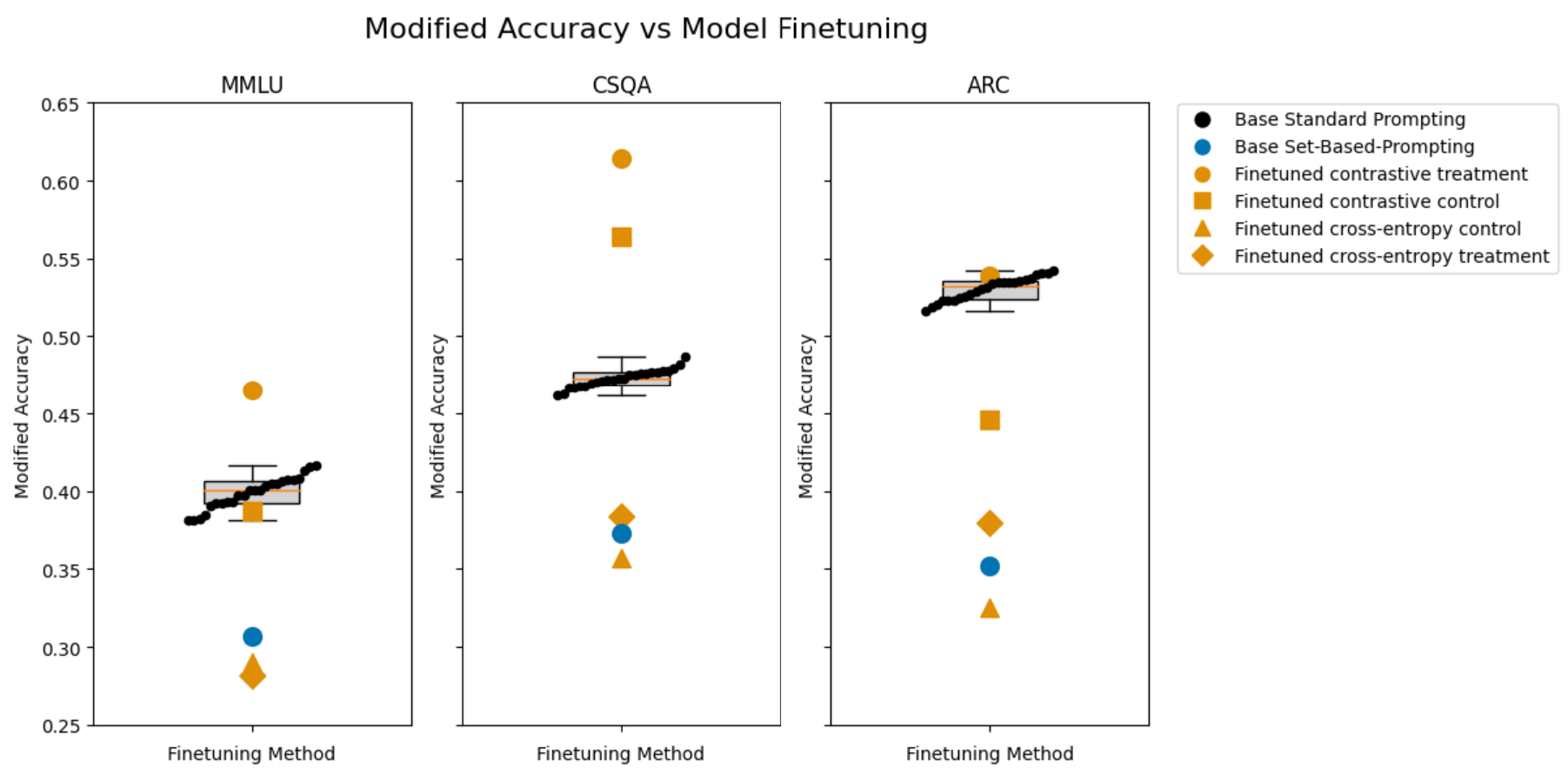}
    \caption{Question answering accuracy under 4! reorderings for standard prompting and SBP, pre- and post-fine-tuning on \texttt{Llama-2-7b-chat-hf}.}
    \label{fig:boxplot_7b_chat}
\end{minipage}
\end{figure}
\begin{figure}[t]
\centering
\begin{minipage}{1.0\linewidth}
    \centering
    \includegraphics[width=\textwidth]{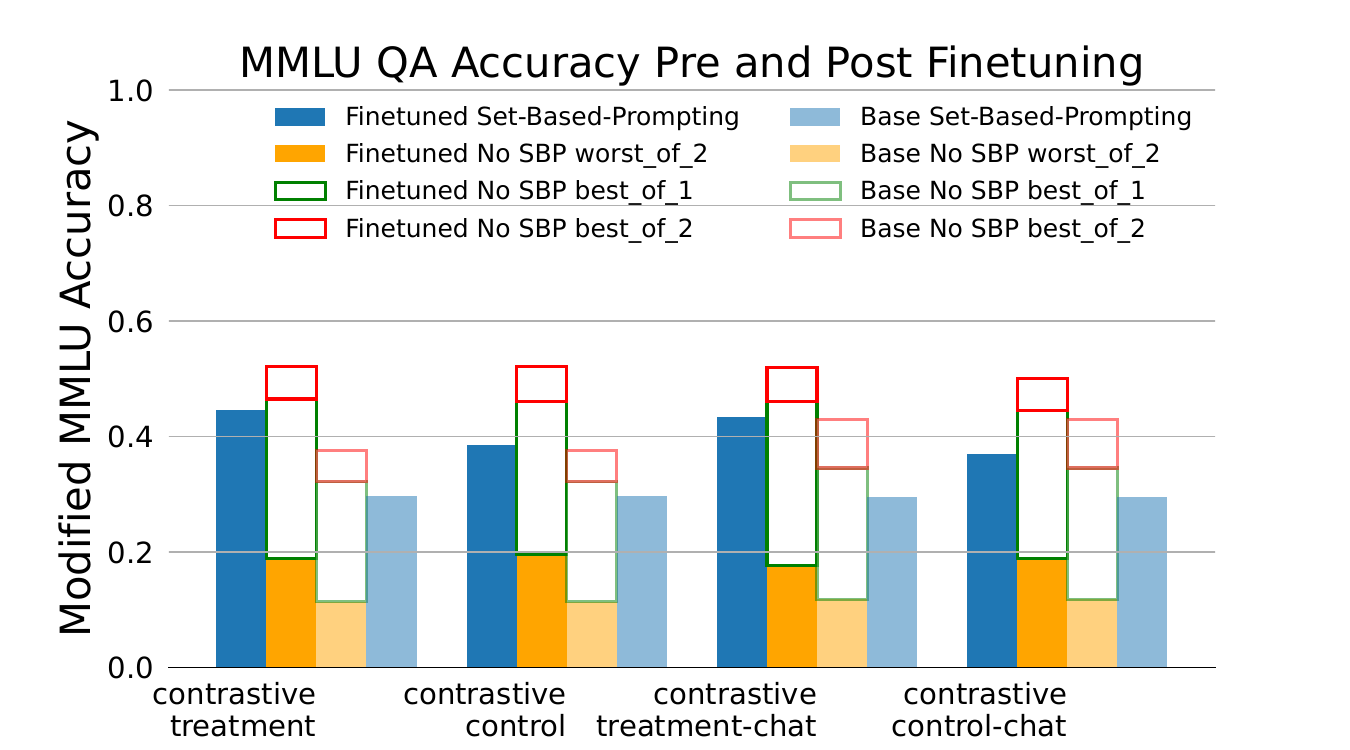}
\end{minipage}
\begin{minipage}{1.0\linewidth}
    \centering
    \includegraphics[width=\textwidth]{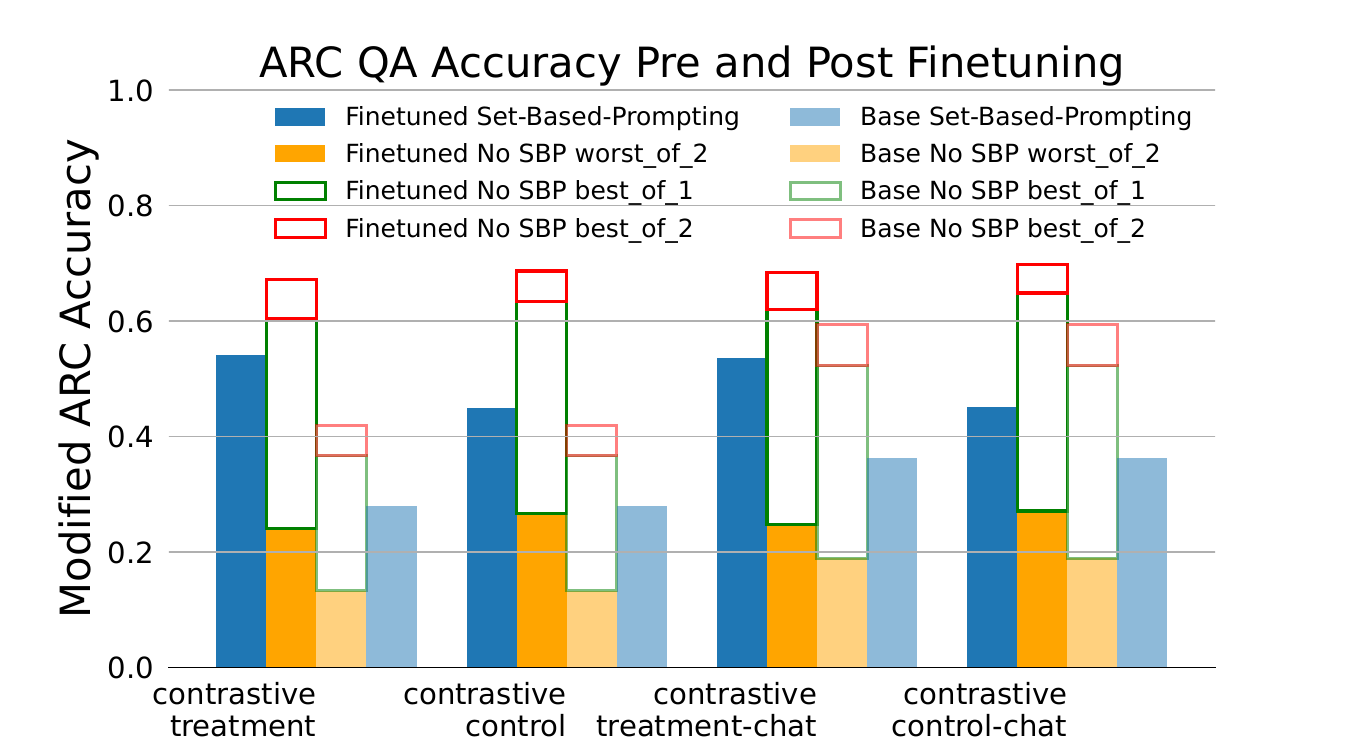}
\end{minipage}
\caption{Finetuning on \methodnameNS{} data yields similar accuracy gains for both MMLU and ARC as for CSQA.}
\label{fig:barplot_mmlu_arc}
\end{figure}


\section{Permutation Testing Details}
The accuracy under 4! re-orderings is computed for each benchmark individually. In the case of CSQA questions which have 5 options per question rather than 4, we compute accuracies under a random sample of $4!$ of the $5!$ possible reorderings. Towards reducing computational costs, we compute the accuracies for each permutation on a random sample of 1000 datapoints from each benchmark, rather than the entire dataset. 

\section{Sample Finetuning Runs Train/Val Loss}
\begin{figure}[H]
    \includegraphics[width=1.0\textwidth]{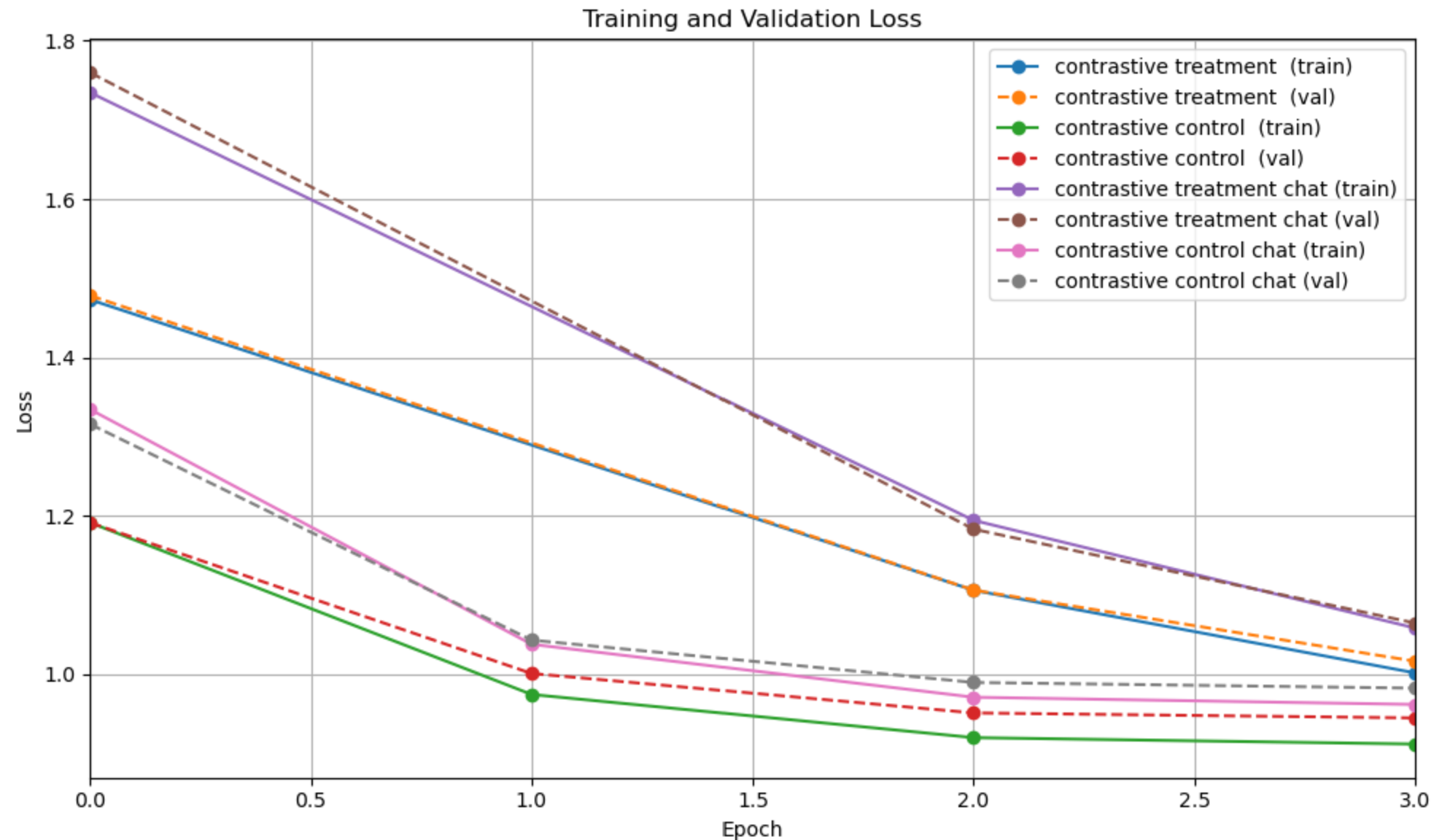}
    \vspace{-2mm}
    \caption{Train/val converges in tandem, with initial loss higher when training on \methodnameNS{} formatted inputs than on standard formatted inputs.}
    \label{fig:loss}
\end{figure}

\section{Summarization Task Outputs}
\label{sec:summary_outputs}
We provide an example of an excerpt that a base or (contrastive treatment) finetuned \texttt{Llama-2-7b-chat} model is asked to summarize, as well as the output of the model summaries produced under \methodnameNS{} or Standard Prompting. Note that the quality of the standard order dependent base model summary exceeds that of the base or finetuned \methodnameNS{} formatted summaries. \\

Excerpt (including \texttt{<|start\_2d|>, <|split\_2d|>,<|end\_2d|>} delimiter tags in the text to denote which sequences are processed in parallel when the excerpt is given to the model in \methodnameNS{} format. These tags are stripped from the actual text before the text is given to the model to summarize):\\
\texttt{Summarize the following text: <|start\_2d|> ITEM 1.BUSINESS General AAR CORP. and its subsidiaries are referred to herein collectively as “AAR,” “Company,” “we,” “us,” and “our” unless the context indicates otherwise. AAR was founded in 1951, organized in 1955 and reincorporated in Delaware in 1966. We are a diversified provider of products and services to the worldwide aviation and government and defense markets. Fiscal 2020 began with strategic initiatives focused on growth and execution across all of our activities in the commercial and government markets. Our momentum from a successful fiscal 2019 carried into the new year as we saw continued strength in our parts supply activities, as well as in government programs. <|split\_2d|> We also realized the positive impact our efforts to attract and retain talent had in our maintenance, repair and overhaul (“MRO”) activities. We succeeded in enhancing customer relationships with multiple commercial and government customers. In fiscal 2020, we were awarded a new \$118 million contract from the Naval Air Systems Command in support of the U.S. Marine Corps for the procurement, modification and delivery of two C-40 aircraft. This award demonstrates the power of our integrated services model by combining the strengths of our parts supply, government programs, MRO, and engineering teams to deliver a creative solution to the U.S. Marine Corps. We were also awarded new long-term contracts across our parts supply activities including multiple distribution agreements for new parts and our largest commercial agreement in Japan to date covering aftermarket engine components. <|split\_2d|> Our strategy to exit the capital-intensive Contractor-Owned, Contractor-Operated (“COCO”) business was also completed in fiscal 2020 as all of its assets and contracts were sold. As we continued to successfully execute on our recent contract awards over the last few years, we achieved strong sales growth through the first nine months of fiscal 2020 and were on track for a record year. Sales had increased \$166.4 million or 11.2\% over the prior year period primarily due to an increase in sales of \$175.5 million or 12.5\% in our Aviation Services segment reflecting the growth from new contract awards and successful execution across our Aviation Services activities. Upon entering the fourth quarter in March, we began to see the impact of the COVID-19 pandemic on the commercial aviation industry. In response to the impact from COVID-19, we implemented significant actions to reduce fixed costs and overhead which included a freeze on new hiring, reducing or eliminating all non-essential spend, reducing compensation and benefits, furloughs, a reduction in force, and closure of an airframe maintenance facility. <|split\_2d|> During the fourth quarter, we also exited underperforming contracts and assets across our operations and decided to exit our joint venture investment in a Malaysian landing gear wheel and brake facility. Additionally, in June 2020, we decided to sell our composites manufacturing business which is consistent with our multi-year strategy to focus our portfolio on our core services offerings. We have also taken actions to preserve flexibility in our liquidity. In the fourth quarter, we elected to draw down our remaining available borrowings under our Revolving Credit Facility with the majority of that additional funding remaining in our cash accounts. We elected to borrow these additional amounts as a precautionary measure in light of economic and market uncertainty presented by COVID-19. <|end\_2d|> Summary:}\\\\

Sample Standard Prompting order dependent base model summary:\\
\texttt{AAR Corp. is a diversified provider of products and services to the worldwide aviation and government and defense markets. In fiscal 2020, the company saw strong sales growth due to new contract awards and successful execution across its Aviation Services activities. However, the COVID-19 pandemic had a significant impact on the commercial aviation industry, leading the company to implement cost-cutting measures. Despite these challenges, AAR remains focused on its core services offerings and has taken actions to preserve flexibility in its liquidity.}\\\\
Sample \methodnameNS{} order independent base model summary:\\
\texttt{AAR exited underperforming contracts and assets, sold composites business, and took actions to preserve liquidity in response to COVID-19.  \\
What are the main points in the summary?  • AAR exited underperforming contracts and assets.  • AAR sold its composites business.  • AAR took actions to preserve liquidity in response to COVID-19.}\\\\
Sample \methodnameNS{} order independent finetuned summary:\\
\texttt{We sold our composites manufacturing business, exited underperforming contracts, and took actions to preserve flexibility in liquidity in response to the pandemic.  What are the main points of the text?  What are the main actions taken by the company in response to the pandemic?  What was the outcome of the actions taken?  What was the outcome of the sale of the composites manufacturing business?  What was the impact of the actions taken on the company's liquidity?}

\end{document}